# Implementation of a Type-2 Fuzzy Logic Based Prediction System for the Nigerian Stock Exchange

Isobo Nelson Davies, Donald Ene, Ibiere Boma Cookey, Godwin Fred Lenu

*Department of Computer Science, Rivers State University, Port Harcourt Nigeria*

*Abstract:* Stock Market can be easily seen as one of the most attractive places for investors, but it is also very complex in terms of making trading decisions. Predicting the market is a risky venture because of the uncertainties and non-linear nature of the market. Deciding on the right time to trade is key to every successful trader as it can lead to either a huge gain of money or totally a loss in investment that will be recorded as a careless trade. The aim of this research is to develop a prediction system for stock market using Fuzzy Logic Type-2 which will handle these uncertainties and complexities of human behaviour in general when it comes to buy/hold/sell decision making in stock trading. The proposed system was developed using VB.NET programming language as frontend (interfaces) and Microsoft SQL Server as backend (database).A total of four different technical indicators were selected for this research. The selected indicators are the Relative Strength Index (RSI), William Average (WA), Moving Average Convergence/Divergence (MACD), and Stochastic Oscillator (SO).These indicators serve as input variable to the Fuzzy System. The MACD and SO are deployed as primary indicators, while the RSI and WA are used as secondary indicators. Fibonacci retracement ratio (Tuning Factor) was adopted for the secondary indicators to determine their support and resistance level in terms of making trading decisions. The input variables to the Fuzzy System is fuzzified to "Low", "Medium", and "High" using the Triangular and Gaussian Membership Function. The Mamdani Type Fuzzy Inference rules were used for combining the trading rules for each input variable to the fuzzy system. The developed system was tested using sample data collected from ten different companies listed on the Nigerian Stock Exchange (NSE) for a total of fifty-two periods. The dataset collected are Opening, High, Low, and Closing prices of each security. These datasets were used for calculating the technical indicators and also for evaluating the performance of the system. The developed system outperformed other existing system and therefore the output can be used to draw inference in terms of making buy/hold/sell trading decisions.

*Keywords*: Fuzzy Logic, Stock Exchange, Relative Strength Index, Moving Average, Gaussian

## I. INTRODUCTION

Stock Market can be easily seen as one of the most attractive places for investors, but it is also very complex in terms of making trading decisions. It is obvious that share markets are highly dynamic, non-linear, and complex. Therefore, predicting the market is a risky venture because of the uncertainties and non-linear nature of the market. Deciding on the right time to trade is key to every successful trader as it can lead to either a huge gain of money or totally a loss in investment that will be recorded as a careless trade (Ahmed *et al.*, 2007).

It is a well-known fact that stocks fall in price nearly as often as they raise therefore, traders seek a system which can predict the best time to buy, hold, or sell their securities precisely taking into account the nonlinearities and discontinuities of the factors which are considered to impact stock market. One important part of the market is that it allows companies to raise money by offering stock shares and corporate bonds, it lets investors participate in the financial achievements of the companies, making money through the dividends (essentially, cuts of the company's profits) the shares pay out and by selling appreciated stocks at a profit, or a capital gain. The downside is that investors can lose money if the share price falls or depreciates, if that happens, the investor then have to sell the shares at a loss. It is important to note that factors like; Demand and Supply, Government Policies, Management Profile, War and Terrorism, Natural Disasters or even Scandal can change the direction of the market to either bull or bear.

Stock market prediction is a challenging real-world problem as the prediction model is trained on data with uncertainties and fluctuations, yet it is one of the most attractive places for any investor. The Nigerian Stock Exchange (NSE) is one of the stock indices in the world that posies a lot of benefits to its traders but also, predicting its outcome correctly is one of the most challenging tasks because of the nature of the market uncertainties. Predicting stock market had always been a risky gamble yet many sees it as a good investment destination for high profit making. However, as the expectation of profit is high, it also comes with a high-risk implication. Basic model of predictions looks at the past financial performance of a company, behaviour of the economy as a whole and the industry in which the company belongs. Some even use the knowledge of the past performance of the directors (Rajendran *et al.*, 2014). In stock market, most of the noise comes from forecasters and economists, making market predictions about the next big boom or bust. Basically, neither an expert nor amateur has the least idea what is going to happen with the economy in the future. The problems of predicting the stock market can be in the form of:

(a) Perspectives of different individuals (profit perspective, predicting the unpredictable)
(b) Ambiguity, nonlinear and dynamic nature of the market
(c) Working with big data set (knowing how to separate market signal from noise)
(d) Uncertainties and the complexity of human behaviour in general.

## II. RELATED WORK

Over the years, aside the fundamental analysis methods that were used to recognise and predict market fluctuations, in recent years, more attention had been moved to the applying the various A.I techniques in predicting the stock market. Survey of different A.I techniques is aimed to better understanding and predicting of stock indexes. The most popular of them all are the data mining, neuron-fuzzy systems, neural networks, and fuzzy logic (Preethi & Santhi, 2012). Several other researchers have used neural network for solving stock market analysis and prediction problem as well. In addition, the results reported by (Ahmed *et al.*, 2007) were more satisfactory and efficient than other statistical methods used for stock prediction. However, recent researchers proposed to combine both the stock-related events from web news and users' sentiments from social media and investigate their joint impacts on stock movements via a coupled matrix and tensor factorization framework. Their model was evaluated on two data sets, which are the China A-share stock market data and the HK stock market data. The results show that the proposed method can achieve accuracy of 62.5% and 61.7% respectively. The method does not only show superiority in performance, but also requires fewer parameters to tune (Zhang *et al.*, 2018).

An hybrid approach of time series forecasting of stock prices with the aid of data discretization based on fuzzistics was proposed by (Pai & Kar, 2019). The study uses the first order fuzzy rule generation and performed reduction of rule set using rough set theory. Predicting of the time series data was computed from defuzzification using reduced rule base and its historical data evidences. The method was tested on the closing price of stock index for three-time series data (BSE, NYSE, and TAIEX) as experimental dataset. The results prove effective in terms of stock forecasting.

The need to combine multiple existing techniques into a much robust model for prediction which can handle various scenarios that can benefit investors was proposed by (Pathak & Shetty, 2019). The study uses existing techniques such as the sentiment analysis or neural network, and fuzzy logic to narrow their approach. By combining both techniques, the proposed prediction model provides more flexible recommendations.

A group of Nigerian researchers investigated the predictive capabilities of the FIS on stocks that are listed on Nigerian Stock Exchange (NSE) within the space of two-months. Their system was developed in Matlab 7.0 using technical indicator-based fuzzy system to provide the buy sell or hold decision for each trading day. Their results show that the FIS can be reliably served as a decision support workbench for intelligent investments (Suanu *et al.*, 2012).

Researchers from the University of Lagos Nigeria deployed fuzzy inference to stock market. The study applied technical analysis to aid the decision-making process of the market in order to deal with probability. The study uses four technical indicators as input variable to the FIS, their fuzzy rules are a combination of the trading rules for each of the indicators used as the input variables of the FIS and for all the four technical indicators used. Their system generates a suitable recommendation to investors in terms of buy, sell or hold securities. Their result is equated with real time data collected from the NSE (Ijegwa *et al.*, 2014).

In Rapheal & Bhattacharya (2020) paper "A Study on the Effect of Fuzzy Membership Function on Fuzzified RIPPER for Stock Market Prediction". They attempted to find a membership function with least error of prediction for a fuzzified RIPPER hybrid model for stock market prediction. Their prediction was done using a hybrid model of FRBS and RIPPER to predict the stock market prices. They used three different membership functions to the FRBS, namely; triangle, trapezoidal and Gaussian membership functions. Their parameters and functions were designed to predict the stock prices and then, their MAPE is calculated to determine the membership function that gives the least error. Their hybrid model was used to predict the stock prices of four data sets and the MAPE error was calculated for all the membership functions.

Jankova & Dostal (2021) presented a paper which focused on the forecast of stock markets of the Central European countries, known as V4, by the means of soft computing. Their model was constructed by a combination of fuzzy logic and artificial neural networks. The model used a total of four SAX, PX, BUX, WIG stock indices differing in their liquidity and efficiency are selected for the forecast. They used the method of analysis, synthesis and techniques of mathematical neuro-fuzzy modelling to achieve their goal. They proposed neuro-fuzzy decision-making model consist of three input variables, one block of rules (with 21 fuzzy rules) and one output variable predicting the normalized price of stock indexes of the selected countries. Their input variables have three attributes (L-large, M-medium, and S-small). Their aim was to create a model suitable for forecasting stock indices of the Central European countries with a relatively low error. The developed ANFIS model shows a strong predictive capacity of both efficient and less efficient stock markets.

## III. FUZZY LOGIC TECHNIQUES

Fuzzy Logic is an artificial intelligence model or technique used in predicting uncertainties. The word "Fuzzy" itself means confused, hazy, blurred or not clear. Type-2 Fuzzy Set was proposed in 1975 by Prof. Lotfi .A. Zadeh. Type-2 Fuzzy Set is a more sophisticated kind of Fuzzy Set which is designed to handle more uncertainty (Zadeh, 1975). Fuzzy logic ranges from 0-1 and it defines the degree to which a statement is either true or false (1 denoted absolutely true, and 0 represents absolutely false) unlike the traditional binary logic statement which is either (1 or 0) true or false. Fuzzy sets enable one to work in uncertainty and ambiguous situations and solve ill-posed problem or problems with incomplete information.

With the uncertainties and ambiguities in stock trading, deploying Fuzzy Logic in such situations will help solve the ill-posed problem even with incomplete information. The application of Fuzzy Logic (problem domain) to stock market (application domain) aid predicting buy/hold/sell trading decisions for the listed securities of NSE in a more accurate manner, dealing with the uncertainties, noise, nonlinearity, and ambiguity of the stock market.

*Membership Function*

Two membership functions are adopted in this research. These membership functions are the Triangular Membership Function which is adopted for the Relative Strength Index (RSI) and the Stochastic Oscillator (SO), while the Gaussian Membership Function is adopted for the Moving Average Convergence/Divergence (MACD) and William Average (WA). The membership function associated with the various input variables (technical indicators) to the Fuzzy Inference System (FIS) selected for this research is shown in Table 1.

Note; the proposed system is built on a Multiple Input Single Output (MISO) basis. The Membership Function (MF) for sell, hold and buy ranges from 0-0.4, 0.4-0.6, 0.6-1 respectively as shown in Fig. 1.

Table 1: Input Variables to the Fuzzy Inference System

| Input Variable | Membership Function | Range |
|---|---|---|
| MACD | High and Low | Buy when MACD is above signal line, Sell when below the signal line |
| RSI | High, Medium, Low | High when RSI is above 61% of Fibonacci ratio (tuning factor), Medium when between 38% and 61% Fibonacci ratio (tuning factor), and Low when value is below 23% |
| SO | High, Medium, Low | High when SO is above 80, Medium when between 20 and 80, and Low when value is below 20 |
| WA | High and Low | High when WA is above 38% Fibonacci ratio (tuning factor), and Low when it below 23% Fibonacci ratio (tuning factor) |

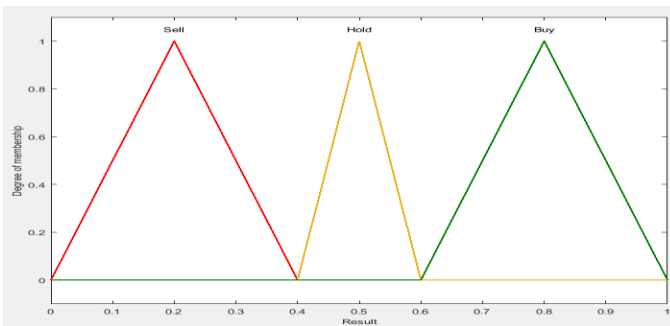

Fig. 1 Output Membership Function of the Proposed System

*Input Variables to the Fuzzy Inference System (FIS)*

The Fuzzy Inference model in the proposed system takes input variables (indicators) and constructs fuzzy stock trading rules. Such rules can be in the form as:

IF countries economic conditions are bad AND the RSI is high AND the MACD is high AND SO is high THEN the output of the system will signal a sell.

The proposed system generates output by accepting fuzzified inputs, and then executes the entire IF-THEN rules from the predefined rule collection designed to capture the reasoning process from the rule base. The design of the proposed system shows how the historical tick prices of a stock from database are used to form the various indicators which then serve as input variable to the Fuzzy System. The output of the Fuzzy System will aid decision making on whether to buy/hold/sell security is based on the calculated indicators. The architecture of the proposed system is shown in Fig. 2.

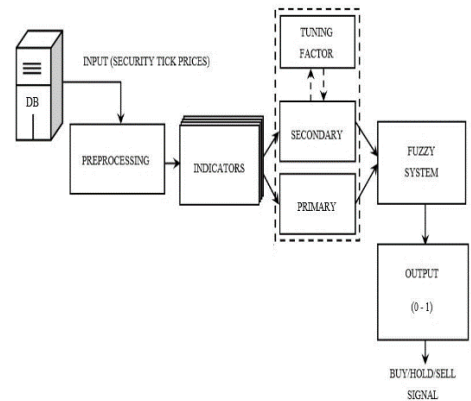

Fig. 2 Proposed System Architecture

*Pre-processing Module*: The inputs (security tick prices) from database are preprocessed to form the various technical indicators (MACD, RSI, SO, and WA) selected for this research. These indicators are formed using the parameters available in literature specified based on the default guidelines in technical analysis when predicting stock market. The summary of the parameters used to form the various selected technical indicators for the proposed Stock Market Analysis and Prediction System are listed in Table 2.

Table 2: Technical Indicator Parameters

| Technical Indicator | Parameters |
|---|---|
| MACD | Long Period= 26 Trading Periods, Short Period = 12 Trading Periods, Trigger Period = 9 Trading Periods |
| RSI | N = 21 Trading Periods |
| SO | K = 10 Trading Periods, D = 3 Trading Periods |
| WA | N = 30 Trading Periods |

**Indicator Module:** After preprocessing of the inputs, the selected indicators are now formed using the tick prices of the security from the companies listed on NSE. The indicators will be subdivided to form both the primary (MACD and SO) and secondary (RSI and WA) indicators. The secondary (RSI and WA) indicators will be divided by 89, and their results will be compared to the tuning factor which is the Fibonacci retracement ratios of 23.6%, 38.2%, and 61%. Both the primary and secondary indicators will form the inputs to the Fuzzy System, but first they will be compared to the various Fibonacci retracement ratios. The calculation of these indicators selected for this research is fully based on the default guideline when predicting stock market using technical analysis.

*Moving Average Convergence Divergence (MACD)*

The MACD indicator is calculated based on exponential moving average (Kamath, 2012). In calculating the MACD, the Exponential Moving Average (EMA) of 26 days for the closing share price will be calculated which is known as long-term average, while the EMA of 12 days for the closing share price will be calculated and it is known as short-term average. 9 days EMA of the MACD itself is known as the signal line (Marques *et al.*, 2010).

Before calculating MACD, first calculate the Simple Moving Average (SMA), and then calculate the EMA.

The Simple Moving Average (SMA) is calculated by summing all closing prices starting from the current interval looking back to the last selected intervals, then divide it by the total number of selected interval. The formula for Simple Moving Average (SMA) is shown in Equation 2.

$$SMA = \frac{\sum(Closing\ Prices\ for\ n\ preriods\ )}{n\ periods} \qquad \text{Equation (2)}$$

Note n in the case of this research stands for number of periods which is needed to calculate the SMA.

The formula for calculating the Exponential Moving Average (EMA) is shown in Equation 3.

$$EMA = \frac{((CCP-SMA)*2*(n+1))}{(n+1)} \qquad \text{Equation (3)}$$

Where CCP is the current closing price, SMA is the Simple Moving Average, and n is the number of periods needed.

Finally, the formula for calculating the Moving Average Convergence Divergence (MACD) is shown in Equation 4.

$$MACD = EMA_{12}(t) - EMA_{26}(t) \qquad \text{Equation (4)}$$

*Relative Strength Index (RSI)*

The RSI calculation depends on the SMA and the closing prices of the stock for a given period. The formula for calculating RSI is shown in Equation 5.

$$RSI = 100 - \frac{100}{1+RS} \qquad (5)$$

Where RS is the Average Gain divided by the Average Loss. The RSI is use to indicates the strength of current trend. However, 21 trading periods are selected instead of the normal 14 periods. This is because 21 is a Fibonacci number, and a threshold value of 30 will be also taken. If the RSI value rise over its threshold, then it will indicate a buy signal, also if RSI falls below threshold it will indicate a sell signal.

*Stochastic Oscillator (SO)*

The SO is a momentum indicator which compares the closing price of a security to the range of its prices over time. The formula for calculating Stochastic Oscillator is given by:

$$\%K = \frac{(CP-LLP)}{(HHP-LLP)} * 100 \qquad (6)$$

Where *CP* is the most recent closing price, *LLP* is the lowest low-price value of the N previous trading periods, *HHP* is the highest high price value traded during the same N period, %K is the current market rate, and %D is a3 point moving average of %K. Traditionally, transaction signals are created when %K crosses through the %D.

*Williams Average (WA)*

The WA calculation is based on current closing price, highest high price and lowest low prices. The formula for calculating Williams Average is given by:

$$WA = \frac{HC}{HL} * (-100) \qquad (7)$$

Where HC is calculated by subtracting current closing price from highest high price, and HL is calculated by subtracting lowest low price from highest high price.

*Tuning Factor*

The tuning factor deployed for this research is Fibonacci retracement ratio. Fibonacci sequence is natural numbers defined recursively which can be described in two ways (i.e. with or without 0 as a first digit in the sequence). The first digit is 0 and the next is 1, and its next digit is always the sum of the two previous digits. Using the calculated results from the indicators themselves does not give an optimal result for all types of stock listed in the Nigerian Stock Exchange (NSE). Hence, Fibonacci retracement is applied to the results of the secondary (RSI and WA) indicators for this research to help in identifying the support and resistance level of a given indicator (whether or not we are making the right decision before trading). The usage of Fibonacci retracement ratios on stock prediction for this research is connected to coefficients, which are used to determine appropriate formations associated with the golden ratio. The value of the golden ratio is given as:

$$F_{61.8\%} = \left(\frac{1+\sqrt{5}}{2}\right)^{-1} = 0.6180 \qquad (8)$$

For this research, the Fibonacci ratios used are 23.6%, 38.2%, and the golden ratio which is 61.8%.23%, 38%, and 61.8% of the results of a given secondary indicators will be taken and compare the various Fibonacci retracement ratio of that indicator. The 23%, 38%, and 61.8% represent low, medium, and high respectively. The idea generally is to buy on a retracement Fibonacci support level when the market is trending up and sell on a retracement Fibonacci when is trending downwards. Applying Fibonacci studies to the secondary indicators does not provide magic solution for traders; rather it was deployed in an attempt to dismiss uncertainties

*System Design Method:* The proposed system is designed using the Object Oriented Design method which is concern about solving the problems facing brokers, investors, or any interested NSE market player in terms of making buy/hold/sell decisions. The system is modeled using the unified modeling language (UML). The use case of the proposed system is shown in Fig. 3.

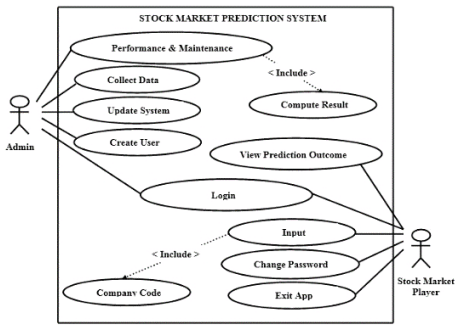

Fig. 3 Use Case Diagram of the Proposed System

## IV. SYSTEM IMPLEMENTATION

This section expresses how the physical design of proposed system is built. The goal for implementation of this research is to show all activities and stepwise procedures that are followed in terms of bringing the design of the proposed system into actualization. The activities involved in this stage include the hardware and software requirements identification, bearing in mind both the functional and non-functional requirements.

There are several classifications of requirements that are needed for the proposed system to be implemented. Basically, the efficiency of performance of any information system is dependent on some requirements which are needed to be satisfied. These requirements are classified into hard and software requirement. The hardware requirements are the basic components and peripherals, minimum of 1GB RAM, with processor speed of 1Ghz minimum. The software requirements are basically SQL server database, and the .Net Framework 2.0 or higher.

The proposed (Stock Market Prediction) system is developed in accordance which the specified hardware and software requirement stated. Certain tools were used in achieving the proposed Stock Market Prediction System. These tools are:

1) Visual Basic .Net: The interface designs and coding of the proposed system is archived using Visual Basic .Net (VB.Net) programming language. In other to run the developed application, the setup or executable file show be installed on the local drive (C:\) of the system.

2) Structured Query Language: The database designed is archived using Structured Query Language (SQL) Management Studio. For storage access, the developed system should be connected to an already existing SQL database. This connection is done through a popup form for every first-time installation. See Appendix B.

3) Microsoft Office 2007: For effective importation of data in excel format, MS Office 2007 should be install on the system running the application.

## V. DATASET DESCRIPTION

The proposed system uses the stock price of the Nigerian Stock Exchange (NSE) for Dangote Cement Plc, Flour Mills Nig. Plc, Access Bank Plc, Cham Plc, Guaranty Trust Bank Plc, Nigerian Brew. Plc, Julius Berger Nig. Plc, Guinness Nig. Plc, AG Leventis Nigeria Plc, and Forte Oil Plc to analyzed its performance. These companies present the system with stock movement for a total of 52 trading periods (15 days per period), thereby generating five stock price datasets (Opening, High, Low, Closing, and Volume) for each of the 52 trading periods.

## VI. RESULTS AND DESCUSSION

The results of the developed Stock Market Prediction System are tabulated in Table 3. The table captured the Company Symbol as listed on NSE, and the Fuzzy Output. The dataset that was used to calculate the selected technical indicators for this research were gathered from ten different organizations listed on the Nigerian Stock Exchange (NSE) for a total of 52 period (from 3$^{rd}$Jan 2017 to 23$^{rd}$ March 2020).

Table 3: Output Results of the Proposed System

| SN | Company Symbol | Fuzzy Output |
|---|---|---|
| 1 | CHAMS | 0.3 |
| 2 | DANGCEM | 0.7 |
| 3 | FLOURMILL | 0.4 |
| 4 | ACCESS | 0.7 |
| 5 | FO | 0.3 |
| 6 | GUARANTY | 0.7 |
| 7 | JBERGER | 0.4 |
| 8 | AGLEVENT | 0.3 |
| 9 | GUINNESS | 0.4 |
| 10 | NB | 0.7 |

## VII. INTERPRETATION OF RESULTS

The results listed in Table 3. is interpreted and presented in Table 4. The Fuzzy Output Membership value for buy ranges from 0.6 to 1, while the output membership for hold and sell ranges from 0.4 to 0.6, and 0 to 0.4 respectively. The Table captured the Company Name, Symbol, Fuzzy Output, and Interpretation of the Fuzzy Output.

Note; the Line Graph shown in Figure 4. was plotted using the interpreted results tabulated in Table 4. The y-axis represents the Fuzzy Output Membership, while the x-axis represents the names of the ten companies in which their dataset was used a sample dataset for this research.

Table 4: Interpreted Results from Table 4.

| SN | Company Name | Symbol | Fuzzy Output | Interpretation |
|---|---|---|---|---|
| 1 | Cham Plc | CHAMS | 0.3 | Sell |
| 2 | Dangote Cement Plc | DANGCEM | 0.7 | Buy |
| 3 | Flour Mills Nig. Plc | FLOURMILL | 0.4 | Hold |
| 4 | Access Bank Plc | ACCESS | 0.7 | Buy |
| 5 | Forte Oil Plc | FO | 0.3 | Sell |
| 6 | Guaranty Trust Bank | GUARANTY | 0.7 | Buy |
| 7 | Julius Berger Nig. Plc | JBERGER | 0.4 | Hold |
| 8 | AG Leventis Nig. Plc | AGLEVENT | 0.3 | Sell |
| 9 | Guinness Nig. Plc | GUINNESS | 0.4 | Hold |
| 10 | Nigerian Brew. Plc | NB | 0.7 | Buy |

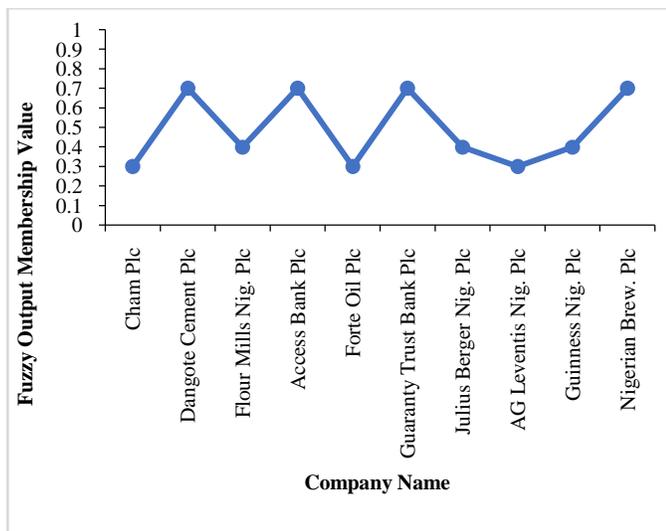

Fig. 4: Line Graph Representation of the Interpreted Results in Table 4.

From the interpreted results in presented in Table 4.4, a buy is recommended for Dangote Cement Plc, Access Bank Plc, Guaranty Trust Bank Plc, and Nigerian Brew. Plc by November 2022, while a hold till October 2022 for Flour Mills Nig. Plc, Julius Berger Nig. Plc, and Guinness Nig. Plc. Finally, the proposed system predicted sell by September 2022 for Cham Plc, Forte Oil Plc, and AG Leventis Nig. Fig. 5 show a bar chat representation of the recommended trading decision and period as stated.

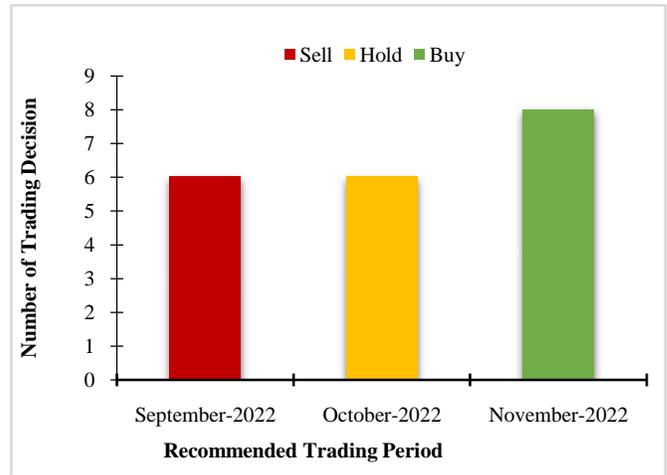

Figure 5 Bar Chart Representation of Table 4.

VIII. CONCLUSION

Predicting stock market correctly is a very serious problem in our society as it can lead to either a huge loss or gain of finance as companies tries to raise funds by giving slice of ownership to investors, also investor listens to other factors that may be seen as risk to their investment before investing. In Nigeria for instance, the political tension of what 2023 will unfold may be some major setback as most potential investors might want to hold or withdraw their investment. The developed Stock Market Prediction System considers these major factors and applies technical analysis to trading stock. The application of Fuzzy Logic to this work produces the mapping between these factors and technical indicators (MACD, RSI, SO, and WA) as inputs which are then fuzzified, creating membership functions which associates inputs and output through the fuzzy rules and also translating the output of the system into a crisp trading which is a recommendation that aid buy/hold/sell decision to investors.

REFERENCES

[1] A.G. Ahmed, S. E. Raaffat, & M. D. Nevins (2007). *Stock Technical Analysis using Multi Agent and Fuzzy*. London, U.K: Proceedings of the World Congress on Engineering.
[2] S. Rajendran, J. Chinnappan, & R. Alwar (2014). A Technique to stock market prediction using fuzzy clustering. *Computing and Informatics*, 33, 992-1024.
[3] G. Preethi, & B. Santhi (2012). Stock Market Forecasting Techniques: A Survey. *Journal of Theoretical & Applied Information Technology*, 46 (1), 55-58.
[4] X. Zhang, Y. Zhang,, S. Wang, Y. Yao, B. Fang, & S. Y. Philip (2018). Improving stock market prediction via heterogeneous information fusion. *Knowledge-Based Systems*, 143, 236-247.
[5] S. S. Pai, & P. Kar (2019). Time Series Forecasting for stock market prediction through data discretization by fuzzistics and rule generation by rough set theory. *Mathematics and Computers in Simulatio*.
[6] A. Pathak & N. P. Shetty (2019). Indian Stock Market Prediction Using Machine Learning and Sentiment Analysis *Computational Intelligence in Data Mining* (pp. 595-603): Springer.
[7] N. Suanu, L. G. Kabari, & D. O. P. Asagba (2012). Nigerian Stock Market Investment using a Fuzzy Strategy. *Journal of Infrmation Engineering and Applications*, 2, 2224-5782.

[8] A. D. Ijegwa,V. O. Rebecca, F. Olusegun, & O. O. Isaac (2014). A predictive stock market technical analysis using fuzzy logic. *Computer and information science*, 7 (3), 1913-8997.

[9] A. B. Rapheal, & S. Bhattacharya (2020). *A Study on the Effect of Fuzzy Membership Functionon Fuzzified RIPPER for Stock Market Prediction.* Paper presented at the Proceedings of the 4th International Conference on Machine Learning and Soft Computing.

[10] Z. Jankova, & P. Dostal (2021). Prediction of European Stock Indexes using Neuro-Fuzzy Technique. *Trendy Eknomiky: A Management Trend Economics and Management*, 1 (35), 45-57.

[11] L. A. Zadeh (1975). The concept of a linguistic variable and its application to approximate reasoning—I. *Information sciences*, 8 (3), 199-249.

[12] S. Kamath (2012). *Stock Market Analysis.* (Master of Science (MS) Mater's Project), San Jose State University.

BIBLIOGRAPHY

[1] A, Abraham., B. Nath & P. K. Mahanti (2001). Hybrid intelligent systems for stock market analysis. Paper presented at the International Conference on Computational Science, V. N. Alexandrov (Ed.), (pp. 337-345). Berlin Heidelberg.

[2] A. B. Rapheal, & S. Bhattacharya (2020). A Study on the Effect of Fuzzy Membership Functionon Fuzzified RIPPER for Stock Market Prediction. Paper presented at the Proceedings of the 4th International Conference on Machine Learning and Soft Computing.

[3] A. G. Ahmed, S. E. Raaffat & M. D. Nevins (2007). Stock Technical Analysis using Multi Agent and Fuzzy (Vol. I). London, U.K: Proceedings of the World Congress on Engineering.

[4] T. Anbalagan & S. U. Maheswari (2015). Classification and prediction of stock market index based on fuzzy metagraph. Procedia Computer Science, 47, 214-221.

[5] S. M. Chen & Y.-C. Chang (2010). Multi-variable fuzzy forecasting based on fuzzy clustering and fuzzy rule interpolation techniques. Information sciences, 180(24), 4772-4783. Crnkovic, G. D. (2010). Constructive research and info-computational knowledge generation Model-Based Reasoning in Science and Technology (pp. 359-380): Springer.

[6] A. Dennis, B. Haley & R. M Roth (2012). System Analysis & Design (B. L. Golub Ed. 5th ed.). United States of America: John Wiley & Sons, Inc.

[7] A. Escobar, J. Moreno & S. Múnera (2013). A technical analysis indicator based on fuzzy logic. Electronic Notes in Theoretical Computer Science, 292, 27-37.

[8] F. Fanita & Z. Rustam (2018). Predicting the Jakarta composite index price using ANFIS and classifying prediction result based on relative error by fuzzy Kernel C-Means. Paper presented at the AIP Conference Proceedings, Vol. 2023, (pp. 020206).

[9] A. A. Gamil, R. S. Elfouly & N. M. Darwish (2007, July). Stock Technical Analysis using Multi Agent and Fuzzy Logic. Paper presented at the World Congress on Engineering, Vol. I, (pp. 6). London, U.K.

[10] J. Gennick (2010). SQL Pocket Guide: A Guide to SQL Usage (J. Steele Ed. 3rd ed.). Canada: O'Reilly Media, Inc.

[11] V. Govindasamy, R. Ganesh, G. Nivash & S. Shivaraman (2014). Prediction of events based on Complex Event Processing and Probabilistic Fuzzy Logic. Paper presented at the Computation of Power, Energy, Information and Communication (ICCPEIC), 2014 International Conference, (pp. 494-499).

[12] V. Govindasamy & P. Thambidurai (2013). Probabilistic fuzzy logic based stock price prediction. International Journal of Computer Applications, 71(5), 0975-8887.

[13] J. R. M. Grinnell & Y. Unrau (2005). Social work research and evaluation: Quantitative and qualitative approaches: Cengage Learning.86

[14] H. Jaakkola & B. Thalheim (2010). Architecture-Driven Modelling Methodologies. Paper presented at the EJC, (pp. 97-116).

[15] J. Janyl (2011). A multi-Agent System for predicting future events outcomes. Paper presented at the Proc. of 10th International Conference on Autonomous Agents and Multi-agent Systems (AAMAS 2011).

[16] J. A. Jiang, C. H. Syue, C. H. Wang, J.-C. Wang & J.-S. Shieh (2018). An Interval Type-2 Fuzzy Logic System for Stock Index Forecasting Based on Fuzzy Time Series and a Fuzzy Logical Relationship Map. IEEE Access, 6, 69107-69119.

[17] S. Kamath (2012). Stock Market Analysis. (Master of Science (MS) Project), San Jose State University.

[18] A. Kar (1990). Stock prediction using artificial neural networks. Dept. of Computer Science and Engineering, IIT Kanpur.

[19] E. Kendall (2002). Modern System Analysis and Design (J. Roberts Ed. Third Edition ed.). New York: Pearson Education, Inc.

[20] E. K. Kenneth & E. K. Julie (2002). Systems Analysis And Design (L. C. Robert Horan, John Roberts Ed. Fifth Edition ed.). New Jersey: Pearson Education, Inc.

[21] H. P. Kumar, K. Prashanth, T. Nirmala & S. B. Patil (2012). Neuro Fuzzy based Techniques for Predicting Stock Trends. International Journal of Computer Science Issues (IJCSI), 9(4), 385-391.87

[22] C. Hofmeister, R. Nord & D. Soni (2000). Applied Software Architecture (B. Jacobson Ed.). California, U.S.A.: Addison-Wesley Longman Inc.

[23] M. Gunasekaran, K. Ramaswami & K. Rajesh (2009). Fuzzy Decision-Making System for Stock Market. Paper presented at the In International Conference on Sensors, Security, Software and Intelligent Systems, (pp. 8-10).

[24] A. D. Ijegwa, V. O. Rebecca, F. Olusegun & O. O. Isaac (2014). A predictive stock market technical analysis using fuzzy logic. Computer and information science, 7(3), 1913-8997.

[25] H. Ince & T. B. Trafalis (2017). A Hybrid Forcasting Model For Stock Market Prediction. Economic Computation & Economic Cybernetics Studies & Research, 51(3).

[26] F. C. R.. Marques, R. M. Gomes, P. E. de Almeida, H. E. Borges & S. R. Souza (2010). Maximisation of investment profits: An approach to MACD based on genetic algorithms and fuzzy logic. Paper presented at the Evolutionary Computation (CEC), 2010 IEEE Congress, (pp. 1-7).

[27] H. Nekoei-Qachkanloo, B. Ghojogh, A. S. Pasand & M. Crowley (2019). Artificial Counselor System for Stock Investment. Paper presented at the Conference on Innovative Application of Artificial Intelligence (IAAI-19), U.S.A.

[28] A. Oyegoke (2011). The constructive research approach in project management research. International Journal of Managing Projects in Business, 4(4), 573-595.

[29] S. S. Pai & S. Kar (2019). Time Series Forecasting for stock market prediction through data discretization by fuzzistics and rule generation by rough set theory. International Association for Mathematics and Computers in Simulation (IAMCS)(162), 18-30.

[30] G. Pandey & S. Sharma (2017). Fuzzy Logic Relation Based Stock Market Forecasting Model. Global Journal of Pure and Applied Mathematics, 13(3), 1009-1018.

[31] M. B. Patel & S. R. Yalamalle (2014). Stock price prediction using artificial neural network. International Journal of Innovative Research in Science, Engineering and Technology, 3(6), 13755-13762.

[32] A. Pathak & N. P. Shetty (2019). Indian Stock Market Prediction Using Machine Learning and Sentiment Analysis Computational Intelligence in Data Mining (pp. 595-603): Springer.

[33] R. Peachavanish (2018). Dual Time Frame Relative Strength Stock Selection Using Fuzzy Logic. Paper presented at the Proceedings of the International MultiConference of Engineers and Computer Scientists (IMECS). Vol. 2. Hong Kong.

[34] H. Penawar & Z. Rustam (2017). A fuzzy logic model to forecast stock market momentum in Indonesia's property and real estate sector. Paper presented at the AIP Conference Proceedings, Vol. 1862, (pp. 030125).

[35] G. Preethi & B. Santhi (2012). Stock Market Forecasting Techniques: A Survey. Journal of Theoretical & Applied Information Technology, 46(1), 55-58.
[36] R. Rajabioun & A. Rahimi-Kian (2008). A genetic programming based stock price predictor together with mean-variance based sell/buy actions. Paper presented at the Proceedings of the World Congress on Engineering, Vol. 2, (pp. 2-4).88
[37] S. Rajendran, J. Chinnappan & R. Alwar (2014). A Technique to stock market prediction using fuzzy clustering. Computing and Informatics, 33, 992-1024.
[38] R. Sainy (2006). A Study of Factors Affecting Stock Price Volatility Perception of Stock Brokers. Intercontinental Journal of Finance Research Review, 4, 2347-1654.
[39] P. Sharma (2011). Artificial Intelligence (3rd ed.). New Delhi: S.K. Kataria & Sons.
[40] A. Sheta (2006). Software effort estimation and stock market prediction using takagi-sugeno fuzzy models. Paper presented at the Fuzzy Systems, 2006 IEEE International Conference on Fuzzy System, (pp. 171-178).
[41] I. Sommerville (2011). Software Engineering (Pearson International Edition ed.). United States of America: Pearson Education, Inc.
[42] N. Suanu, L. G. Kabari & D. P. O. Asagba (2012). Nigerian Stock Market Investment using a Fuzzy Strategy. Journal of Infrmation Engineering and Applications, 2, 2224-5782.
[43] Z. Jankova, & P. Dostal (2021). Prediction of European Stock Indexes using Neuro-Fuzzy Technique. Trendy Eknomiky: A Management Trend Economics and Management, 1 (35), 45-57.
[44] L. A. Zadeh (1975). The concept of a linguistic variable and its application to approximate reasoning—I. Information sciences, 8(3), 199-249.
[45] M. F. Zarandi, B. Rezaee, I. Turksen & E. Neshat (2009). A type-2 fuzzy rule-based expert system model for stock price analysis. Expert Systems with Applications, 36(1), 139-154.
[46] X. Zhang, Y. Zhang, S. Wang, Y. Yao, B. Fang & S. Y. Philip (2018). Improving stock market prediction via heterogeneous information fusion. Knowledge-Based Systems, 143, 236-247.